\documentclass[sigconf]{acmart}
\makeatletter
\def\@ACM@checkaffil{
    \if@ACM@instpresent\else
    \ClassWarningNoLine{\@classname}{No institution present for an affiliation}%
    \fi
    \if@ACM@citypresent\else
    \ClassWarningNoLine{\@classname}{No city present for an affiliation}%
    \fi
    \if@ACM@countrypresent\else
        \ClassWarningNoLine{\@classname}{No country present for an affiliation}%
    \fi
}
\makeatother

\AtBeginDocument{%
  \providecommand\BibTeX{{%
    \normalfont B\kern-0.5em{\scshape i\kern-0.25em b}\kern-0.8em\TeX}}}

\copyrightyear{2021}
\acmYear{2021}
\setcopyright{acmcopyright}
\acmConference[MM '21]{Proceedings of the 29th ACM
International Conference on Multimedia}{October 20--24, 2021}{Virtual Event, China}
\acmBooktitle{Proceedings of the 29th ACM International Conference on Multimedia
(MM '21), October 20--24, 2021, Virtual Event, China}
\acmPrice{15.00}
\acmDOI{10.1145/3474085.3479211}
\acmISBN{978-1-4503-8651-7/21/10}

\begin{document}
\fancyhead{}
\title{Facial Prior Based First Order Motion Model for Micro-expression Generation}

\author{Yi Zhang}
\authornote{Both authors contributed equally to this research.}
\author{Youjun Zhao}
\authornotemark[1]
\author{Yuhang Wen}
\author{Zixuan Tang}
\author{Xinhua Xu}
\affiliation{%
  \institution{School of Intelligent Systems Engineering, \\
  Sun Yat-sen University}
}

\author{Mengyuan Liu}
\authornote{Corresponding Author.}
\email{nkliuyifang@gmail.com}
\affiliation{%
  \institution{School of Intelligent Systems Engineering, \\ Sun Yat-sen University}
}
\affiliation{%
	\institution{Guangdong Provincial Key Laboratory \\ of Fire Science and Technology}
}

\renewcommand{\shortauthors}{Zhang and Zhao, et al.}

\begin{abstract}
 Spotting facial micro-expression from videos finds various potential applications in fields including clinical diagnosis and interrogation, meanwhile this task is still difficult due to the limited scale of training data. To solve this problem, this paper tries to formulate a new task called micro-expression generation and then presents a strong baseline which combines the first order motion model with facial prior knowledge. Given a target face, we intend to drive the face to generate micro-expression videos according to the motion patterns of source videos. Specifically, our new model involves three modules. First, we extract facial prior features from a region focusing module. Second, we estimate facial motion using key points and local affine transformations with a motion prediction module. Third, expression generation module is used to drive the target face to generate videos. We train our model on public CASME II, SAMM and SMIC datasets and then use the model to generate new micro-expression videos for evaluation. Our model achieves the first place in the Facial Micro-Expression Challenge 2021 (MEGC2021), where our superior performance is verified by three experts with Facial Action Coding System certification. Source code is provided in  \url{https://github.com/Necolizer/Facial-Prior-Based-FOMM}.
\end{abstract}

\ccsdesc[500]{Computing methodologies~Computer vision problems}
\ccsdesc[500]{Applied computing~Psychology}
\ccsdesc[300]{Computing methodologies~Image processing}

\keywords{Micro-expression generation, Facial micro-expression, Generative adversarial network, Deep learning, Facial landmark}


\maketitle

\section{Introduction}
Micro-expressions are brief and involuntary facial expressions that usually last for 1/25 to 1/5 of a second.
It is hard for human being to notice facial micro-expressions (FMEs) due to short duration, low intensity, and local-only occurrence \cite{zhou2021objective,Shen2012}.
Recently, spotting micro-expressions with machine learning methods have attracted lots of attentions.
Data-driving machine learning methods need large scale data to obtain an accurate yet robust model.
However, building high-quality FME datasets would be expensive and time-consuming, which leads to the problems: 1) The size of constructed databases is usually too small to assure the robustness of machine learning models. 2) the imbalanced data distribution of currently available databases may lead to unsatisfying training results \cite{10.1007/978-3-319-16817-3_3}. 

To solve this problem, one solution is facial micro-expressions (FME) generation.
Conceptually, the goal of FME generation is to generate micro-expression on given template faces. Challenges in FME generation task are the follows: 1) Few references since it just started studying recently. 2) FMEs are subtle and hard-to-capture. Algorithms often have problem obtaining information of small changes and motions. 3) Traditional image processing methods utilizing handcrafted features get stuck in generation tasks, since the visual result of reconstruction is usually bad, and these methods have poor versatility. Deep learning methods need to be applied.

This paper presents a new method named Facial Prior Based First Order Motion Model for micro-expression generation, where we first extract the motion patterns using the regions feature computed according to the facial prior, and then utilize generative adversarial network to reconstruct the face and generate video. Visual results are involved to show the superior performance of our method. Also, evaluations from three experts verified that our method outperforms other methods in MEGC2021.

\section{Related Work}
Image animation is a way to generate videos by animating objects in still images. Most popular ways of image animation is through deep learning, such as Generative Adversarial Networks (GANs) \cite{10.5555/2969033.2969125} and Variational Auto-Encoders (VAEs) \cite{kingma2014autoencoding}. However, these methods are only used in generating the macro, exaggerated expressions. Since FMEs are small, these methods may not lead to a convincing result. Apparently, a satisfying result should take the subtle movement of the important facial areas into considerations.

Deep generative models for image animations and video retargeting \cite{Bansal2018,10.1145/3343031.3351020,zablotskaia2019dwnet} have emerged in recent years. Some models have been proposed, such as Monkey-Net \cite{siarohin2019animating}, FOMM \cite{NEURIPS2019_31c0b36a} and MRAA \cite{siarohin2021motion} to get better performance in modeling object’s motion. These models encode motion information of the key points or areas in the videos, which are learned in self-supervised way. The performance of self-supervised depends on the diversity of samples. Concerning the motions of micro-expressions are too subtle for models to capture,  self-supervised module should be replaced with prior knowledge about FMEs so that reconstructed videos could be better. Qiao et al \cite{qiao2018geometrycontrastive} proposed Geometry-Contrastive GAN for Facial Expression Transfer which is related to our work. However, their work is limited to macro-expression transfer.

\section{Problem Formulation}
We have a target image $T$, and a series of source images $S={\{S_k\}}_{k=1}^{n}$, also called a driving video, where $S_1$ represents the onset frame and $S_n$ represents the offset frame of a micro-expression. 
Suppose function $Motion(S_i,S_j)$ is the motion representation from frame $S_i$ to $S_j$, and a function $Move(T_1,Motion(\cdot))$ transforms the target face according to the motion representation using distortion, rotation, or other affine transformation. Then this type of FME generation could be formulated as:
\begin{equation}
T_k=Move(T_i,Motion(S_j,S_k))
\end{equation}

There would be various subclasses if we take different $T_i$ and $S_j$. Like inter-frame motion estimation, which is defined as:
\begin{equation}
T_k=Move(T_{k-1},Motion(S_{k-1},S_k))
\end{equation}

We believe this kind of problem modeling is preferable and our proposed method is based on this methodology. We hope for a model which could generate $T_k$ having the same appearance of the face in S as well as containing the semantic information of motion extracted from the driving frame $S_k$. Let the generation model be $\mathcal{G}$, then the problem can be described as $T_k=\mathcal{G}(T_1,S_k)$. The generated video is $GT=\{T_k\}_{k=1}^{n}=\mathcal{G}(T,S)$.
Two modules are needed in $\mathcal{G}$: motion extraction and face reconstruction, denoted as $\mathcal{M}$ and $\mathcal{R}$ (similar to $Motion(\cdot)$ and $Move(\cdot)$ defined above). Specifically, the reconstruction module should utilize information extracted from the motion extraction module. Then $\mathcal{G}$ could be specified as $\mathcal{G}(\cdot)=\mathcal{R}(\mathcal{M}(\cdot))$.

Since the target face is practically different from driving faces, inspired by FOMM \cite{NEURIPS2019_31c0b36a} and MRAA \cite{siarohin2021motion}, we assume a virtual reference frame $R$ and an operator $\circ$ representing superposition of motions. Then our method can be formulated as:
\begin{equation}
GT=\mathcal{G}(T,S)=\mathcal{R}(T,\mathcal{M}(T,R)\circ\mathcal{M}(R,S))
\end{equation}

\begin{figure}[t]
	\centering
	\includegraphics[width=\linewidth]{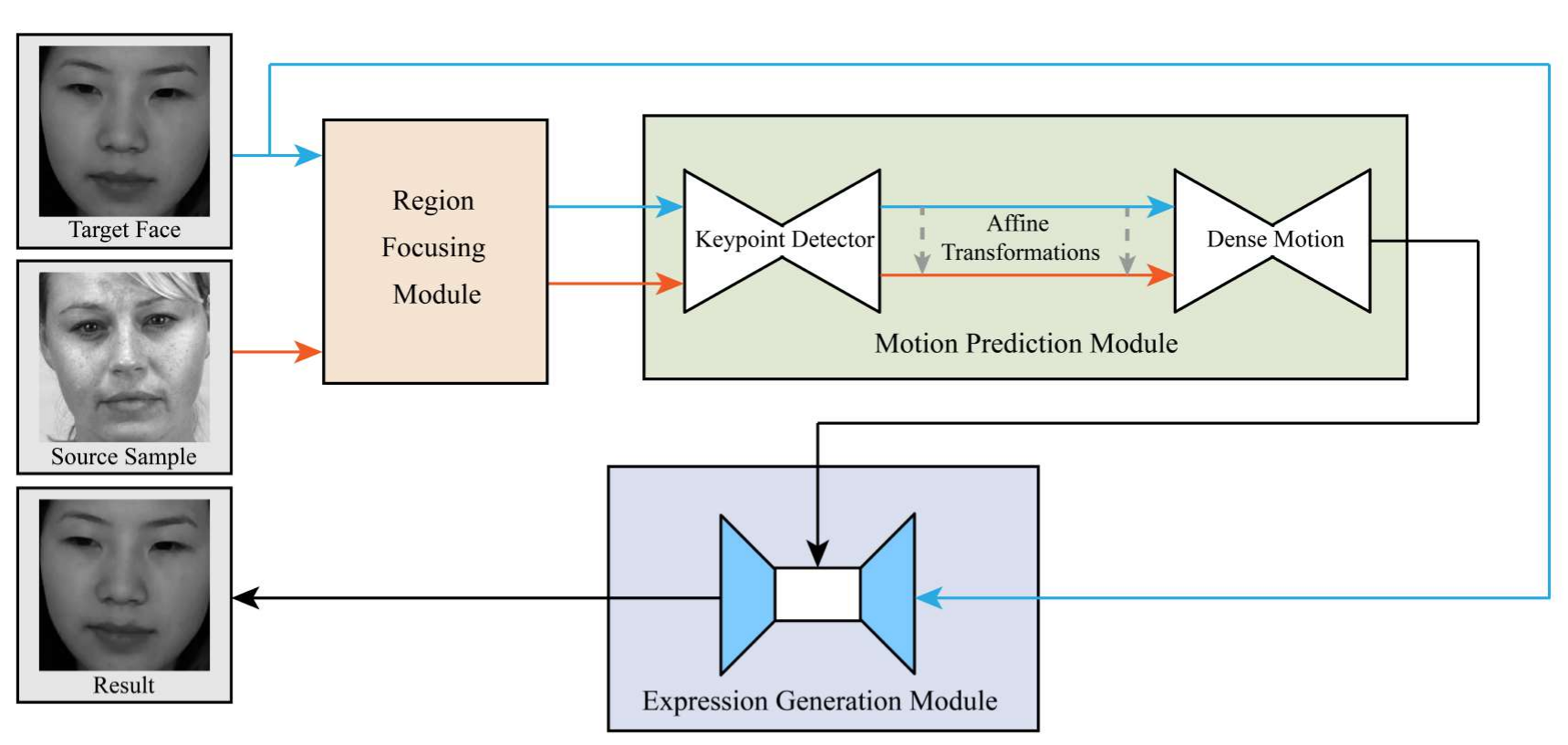}
	\caption{General framework of our method}
	\vspace{-1em}
\end{figure}

\section{Proposed Method}
Fig. 1 shows the general framework of our method. Given a target face, along with a video containing FME, Region-Focusing Module computes a facial prior map. Then the facial prior map is fused with the original frame as the input of Motion Prediction Module. This module uses the key points given explicitly along with local affine transformations to estimate complex motions by predicting backward optical flow. Besides, this module also predicts the occlusion map that indicates the part which could not estimate from the target image, improving the quality of the generated videos. In Expression Generation module, GAN makes use of the predicted optical flow and the occlusion map while encoding so that the decoded image contains the motion information, driving the target to generate FME videos.

\subsection{Region-Focusing Module}
In this section, we will illustrate the algorithm we designed to highlight regions of interest (ROI) for FME.

As mentioned above, there are many priors could be used to extract the feature represented the motion of the driving videos. We noticed that the Facial Action Coding Systems(FACs), which displays 68 facial landmarks in human faces, is a good prior to locate the regions in faces. With the pre-trained model in dlib \cite{Kazemi2014}, the 68 facial landmarks can be automatically predicted, as shown in Fig. 2 (a). But we noticed that not all regions are necessary since FME may only appear in some certain regions. Now the problem is how to focus on the ROI for the FME motion prediction, where the most obvious movements may occur. 

We measure the importance of each pixel according to its distance away from the key points mainly selected from the 68 facial landmarks, which are colored in Fig. 2 (b). Some of the key points are moved horizontally and vertically from the facial landmarks by half the pupillary distance. Specifically, we denote the $k^{th}$ key point in a facial image is $p_k\left(x_k,y_k\right)\in\mathbb{R}^{H\times W}$, and the certain $i^{th}$ pixel in the facial image as $p_i\left(x_i,y_i\right)\in\mathbb{R}^{H\times W}$. Their Euclidean distance $d_{ik}$ is calculated as
\begin{equation}
d_{ik}=\sqrt{\left(x_k-x_i\right)^2+\left(y_k-y_i\right)^2}
\end{equation}

With the calculated distance, we could indicate the importance of this pixel using the gaussian kernel function, described as
\begin{equation}
r_{ik}\left(x_i,y_i\right)=e^{-\frac{d_{ik}}{2\sigma^2}}
\end{equation}
in which $e$ is the base of natural logarithm, $\sigma$ is the variance of the gaussian distribution (set manually). Now the $r_{ik}$ indicates the degree of interest of this pixel concerning the key point $p_k$. With each pixel’s degree of interest to the $p_k$, a map $ S_{km}$(Fig. 3 (a)), which indicates weighted importance, is formulated as:
\begin{equation}
S_{km}=\sum_{r_{ik}\in\mathbb{R}^{H\times W}}{r_{ik}\left(x_i,y_i\right)}
\end{equation}

Assume that n key points are selected, then we could get a synthesized facial prior map $S_m$ in Fig. 3 (b), described as $S_m=\sum_{k} S_{km}$. The original frame image is fused with the facial prior map, forming the sample $S^\prime\in\mathbb{R}^{\left(C+1\right)\times H\times W}$. $S^\prime$ more information related to the ROI. Together with the corresponding predicted key points, these manual-feature enhanced samples are next passed to the motion prediction module for further feature representation using neural network.

\begin{figure}[t]
	\centering
	\includegraphics[width=\linewidth]{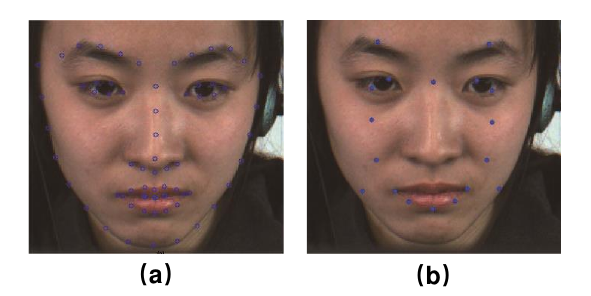}
	\vspace{-2em}
	\caption{Facial prior generation: (a) Original detected key points using facial landmark detection method. (b) Modified key points used as facial prior for our framework.}
	\vspace{-1em}
\end{figure}

\begin{figure}[t]
	\centering
	\includegraphics[width=\linewidth]{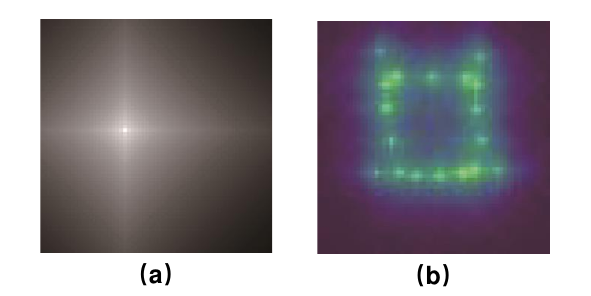}
	\vspace{-2em}
	\caption{Facial prior map generation: (a) Contribution of one key point for the facial prior map. (b) Facial prior map generated with all modified key points.}
	\vspace{-1em}
\end{figure}

\subsection{Motion Prediction Module}
Motion Prediction Module is a two-step model proposed by Siarohin et al \cite{NEURIPS2019_31c0b36a}.
This module can estimate the motion tendency from a driving frame image $S$ in source video to the input target image $T$. The first step is utilizing a key point detector to extract key points in $T$ and $S$ as well as a local affine transformation to represent motion in the local area of each key point from $S$ to $T$. Taylor expansion is used in calculating the affine transformation parameters of key points from $S$ to $T$. Additionally, by introducing reference frame $R$, the prediction of transformation from $S$ to $T$ can be divided into two parts - from $S$ to $R$ and then $R$ to $T$. During the second step, a dense motion network obtains a backward optical flow and a occlusion mask from $S$ to $T$ to decide which part of the image should be reconstructed after getting the transformation parameters from the previous step.

Inspired by this work, we designed Motion Prediction Module as follows. We firstly concatenate image and facial prior map $S_m$ synthesized as an additional channel, instead of putting images directly to the key point detector. Therefore, target matrix and driving matrix of $S^\prime\in\mathbb{R}^{\left(C+1\right)\times H\times W}$ are sent into key point detector to extract features. The additive facial prior map represents the knowledge of occurrence area of micro-expression. With facial prior map, the output key points from key point detector can be located mainly focusing on the neighborhood of micro-expression areas and the movement of the entire head, which can be moved from $S$ to $T$s using the key point trajectories in the subsequent local affine transformation.

\subsection{Expression Generation Module}
Generative adversarial network is taken in our Expression Generation Module. An auto-encoder structure is applied to generate image with a target image $T$ as input. The backward optical flow and occlusive map calculated in {\bfseries4.3} are sent into the encoder block to warp the feature map produced from two layers of down-sampling convolution. Then, a discriminator similar to pixel-to-pixel structure is used to decide if the reconstructed face image $\hat{S}$ from generator is real or not. 

Several losses are applied to train our framework since multiple networks are used in our system. We use an end-to-end perceptual loss of Johnson et al. \cite{Johnson2016} using pre-trained Vgg-19 and Mean Absolute Error (MAE) in other networks. Different weight coefficients are applied to our functions in which the perceptual loss is the highest. Given the reconstructed face image $\hat{S}$ and driving frame $S$, the perceptual loss can be described as:
\begin{equation}
L_{perceptual}\left(\ \hat{S},S\right)=\sum_{i=1}^{I}\left|F_i\ \left(\hat{S}\right)-F_i(S)\right|
\end{equation}
where $F_i$ represents $i^{th}$ channel of the feature map in Vgg-19 and $I$ is the number of feature channels in this layer.

\begin{figure}[t]
	\centering
	\includegraphics[width=\linewidth]{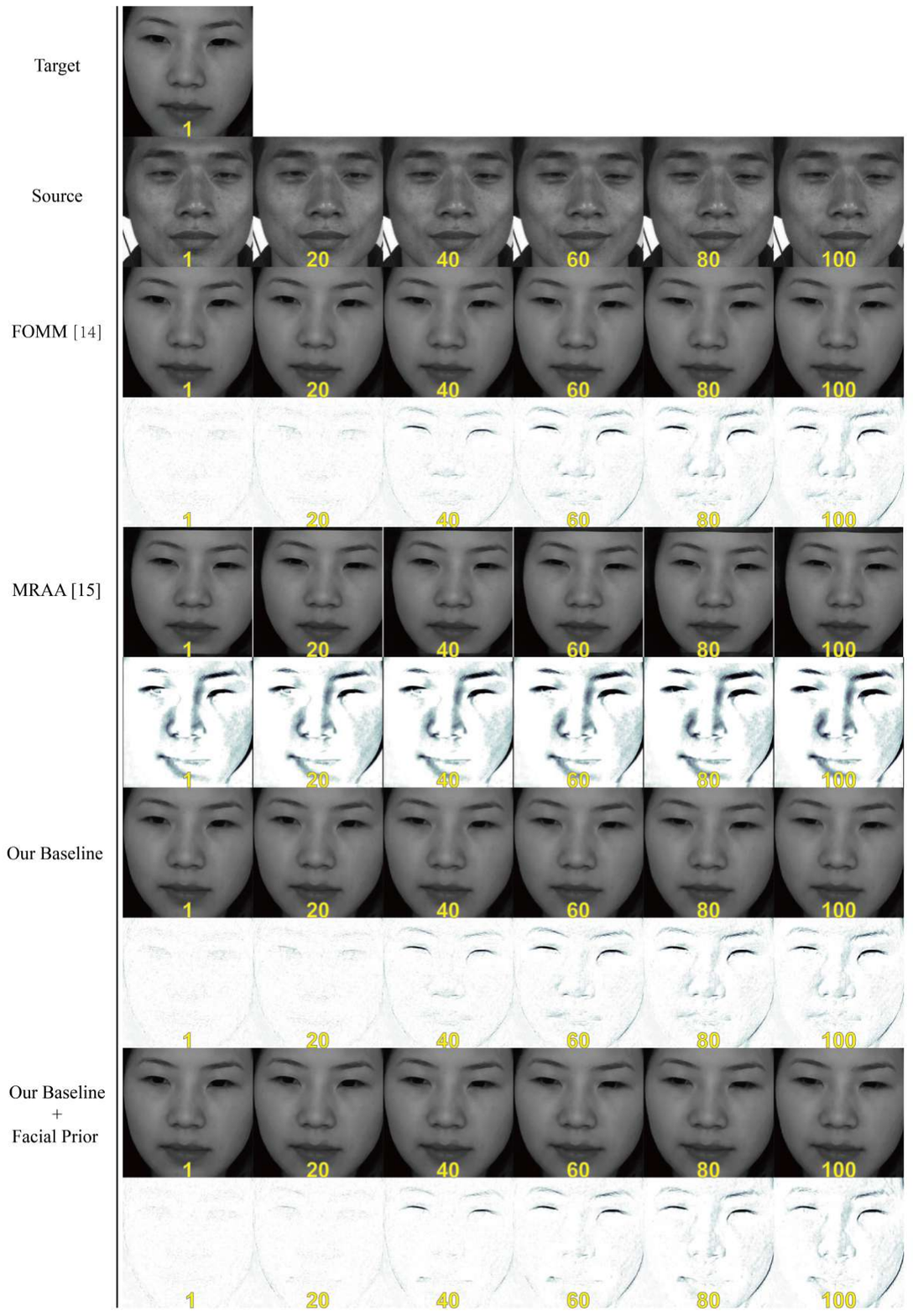}
	\caption{Results of different methods. To facilitate observation, we also show images on the 4th, 6th, 8th, 10th Lines, which are differential images between the 3rd, 5th, 7th, 9th Lines and the target image.}
	\vspace{-2em}
\end{figure}

\begin{figure}[t]
	\centering
	\includegraphics[width=\linewidth]{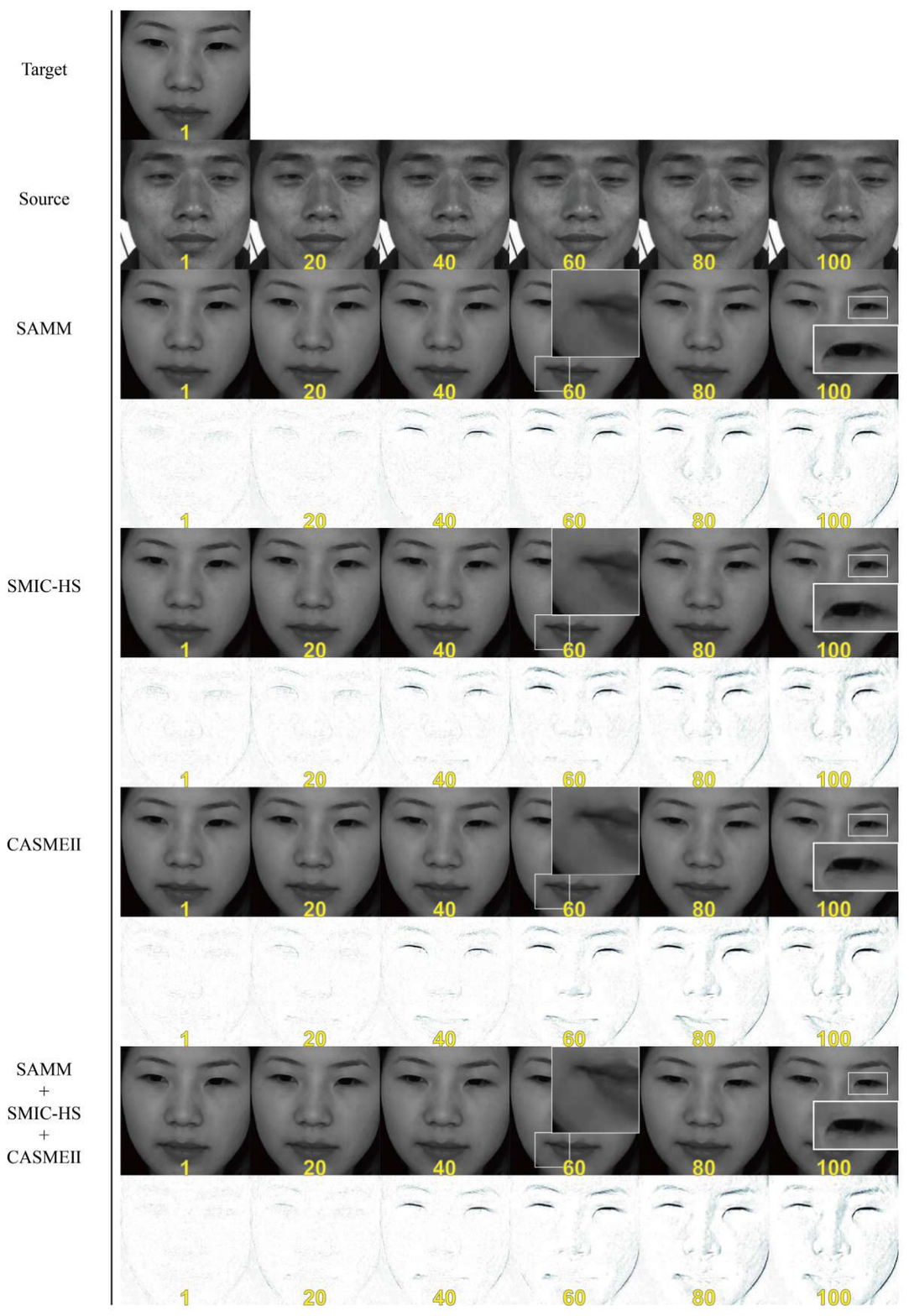}
	\caption{Our results with different training datasets}
	\vspace{-2em}
\end{figure}

\begin{table*}
	\centering
	\caption{Overall evaluation of MEGC2021 Generation Challenge}
	\vspace{-0.5em}
	\label{tab:freq}
	\begin{tabular}{lcccccccc}
		\toprule
		Methods&Expert1&Expert2&Expert3&Overall & \multicolumn{4}{c}{Normalized}\\
		&&&&&Expert1&Expert2&Expert3&Overall\\
		\midrule
		A(ID:3311) & 85 & 51 & 37 & 173 & 85/140 & 51/107 & 37/76 & 1.57062\\
		B(ID:3320) & 104 & 66 & 66 & 236 & 104/140 & 66/107 & 66/76 & 2.228101\\
		C(ID:3282) & 140 & 107 & 56 & 303 & 140/140	& 107/107 & 56/76 & 2.736842\\
		\textbf{Ours(ID:3295)} & \textbf{139} & \textbf{101} & \textbf{76} & \textbf{316} & \textbf{139/140}	& \textbf{101/107} & \textbf{76/76} & \textbf{2.936782}\\
		\bottomrule
	\end{tabular}
\end{table*}

\section{Experiments}
We evaluated our method using CASME II \cite{Yan2014}, SAMM \cite{Davison2018} and SMIC \cite{Li2013} datasets. Given 2 normalized template faces as target (selected from CASME I \cite{Yan2013} and SMIC), we were required to generate FMEs using 9 certain driving videos as source. So we trained our model on the whole datasets except the 9 source videos and generated FMEs on target faces for expert evaluation.
Preprocessing included converting to grayscale, detecting face, cropping the facial area, resizing to $256\times256$, and concatenating the same grayscale to get an image with 3 channels. Our final results are also grayscales because in our opinions generated grayscales seem visually better than color images.
Key points were selected as Fig. 2(b). Every frame in a video shared the same set of key points predicted by dlib according to the onset frame. In this way the computing costs were reduced. In addition, synthesized facial prior map $S_m$ got normalized before entering Motion Prediction Module.

The generation challenge result of MEGC2021, presented in Table 1, shows that our method outperforms methods proposed by other participating teams and won the first place through subjective scoring by three experts who have Facial Action Coding System (FACS) certification \cite{Ekman1978FacialAC}, provided in \url{https://megc2021.github.io/GeneResultevaluation.html}.
Fig. 4. presents the visual results of experiment conducted by ourselves, which compares our proposed method with other baseline methods. The numbers indicate the current frame numbers. We suggest visiting our GitHub page above for GIF to get more intuitive visual results. Compared with FOMM and MRAA, our method could generate more spontaneous, natural, and smooth FMEs with less noise. Also, the effectiveness of facial prior is proved compared with baseline. Moreover, we trained our model on 4 different training sets, which are SAMM, SMIC-HS, CASME II, and a mix of these three. We found that different training datasets would lead to nuances, as shown in Fig. 5.

\section{Conclusion}
This paper presents a facial-prior-based first order motion model for facial micro-expression generation. Specifically, We take prior of the facial area into considerations and design a region-focusing module to get facial prior features, a motion prediction module to estimate facial motions and an expression generation module to generate FME videos. By training on CASME II, SAMM and SMIC datasets, our method achieves superior performances verified by three experts with Facial Action Coding System certification.


\bibliographystyle{ACM-Reference-Format}
\bibliography{FPBFOMM}

\appendix

\end{document}